\documentclass{article} 
\usepackage[final]{colm2026_conference}
\usepackage{microtype}
\usepackage{hyperref}
\usepackage{url}
\usepackage{booktabs}
\usepackage{amsmath}
\usepackage{graphicx}
\usepackage{xcolor}

\usepackage{lineno}
\definecolor{darkblue}{rgb}{0, 0, 0.5}
\hypersetup{colorlinks=true, citecolor=darkblue, linkcolor=darkblue, urlcolor=darkblue}
\title{Human vs Machine Translation Detection: \\ A Cross-Model, Cross-Domain, and Low-Resource Analysis}

\author{%
Yabsera Yemanberhan$^{1}$, Prasenjit Mitra$^{1}$, Roald Eiselen$^{2}$ \\[3pt]
$^{1}$Carnegie Mellon University Africa \qquad $^{2}$Centre for Text Technology, North-West University \\[2pt]
\texttt{\{yhy,prasenjm\}@andrew.cmu.edu} \quad \texttt{Roald.Eiselen@nwu.ac.za}%
}
\begin{document}
\ifcolmsubmission
\linenumbers
\fi
\maketitle
\begin{abstract}
The growing reliance on web-crawled corpora for training large language models introduces a critical challenge: machine-translated (MT) content is often treated as human translation (HT), potentially degrading data quality and propagating synthetic artifacts, especially in low-resource settings. In this work, we study MT--HT detectability across high-resource (English, Spanish) and low-resource (Swahili, Afrikaans) languages using both decoder-only LLMs in zero-shot and few-shot settings, and fine-tuned multilingual encoders trained at the sentence level and on token-chunked inputs. We find that decoder-only models provide limited discrimination overall and exhibit a consistent class-asymmetric failure mode: they recognize human-written text reliably (high HT recall) but rarely flag fluent machine output (low MT recall). Encoder-based classifiers achieve high accuracy when evaluated on data generated by the same MT systems and translators used during training in a legislative domain, but their MT recall drops sharply on unseen translators and on unseen domains such as low-resource conversational text and medical content. To address this limitation, we evaluate models on longer inputs by chunking text into segments of up to 300 tokens, which substantially improves generalization, suggesting that sentence-level signals are insufficient for robust detection. Finally, we analyze the relationship between translation quality and detectability using COMET and BLEURT, showing that lower-quality MT is more easily identified, while high-quality MT remains difficult to distinguish from HT. Our results highlight the importance of input granularity, translator generalization, and translation quality in building robust MT detection systems across domains and languages.
\end{abstract}

\section{Introduction}
Large-scale multilingual corpora such as Common Crawl~\citep{CommonCrawl} and CCAligned~\citep{ElKishky2020CCAligned} underpin the training of modern machine translation (MT) systems and multilingual language models. However, these corpora have been shown to contain machine-generated text, including machine-translated (MT) content that is indistinguishable from human-translated (HT) text~\citep{Dodge2021C4}. This form of \emph{MT contamination} can mislead downstream models into learning synthetic artifacts rather than natural linguistic patterns, with particularly harmful effects in low-resource settings where training data is already scarce. Prior work has shown that MT can alter linguistic richness and introduce systematic biases~\citep{Vanmassenhove2019Lost}, and that MT and HT texts can be distinguished using source-oriented features~\citep{NguyenSon2021TSRT}, but the robustness of such methods across domains, resource levels, and modern MT systems remains unclear.

Detecting MT in web-scale corpora is challenging for two complementary reasons. In high-resource settings, modern neural MT systems produce highly fluent outputs that weaken traditional translationese signals such as lexical repetition or syntactic rigidity~\citep{Aharoni2014Automatic}. Moreover, stylistic convergence across MT systems reduces system-specific artifacts, leaving only subtle discourse-level differences between MT and HT. In low-resource settings, the challenge shifts from weak signals to variability: translation quality varies widely across languages, domains, and systems, leading to inconsistent detection cues and poor generalization. As a result, models trained on specific domains or MT systems often fail to transfer across settings, particularly for low-resource languages and unseen translators.

In this work, we systematically study MT--HT detectability across language-resource conditions, domains, and model families. To our knowledge, this is the first study to systematically evaluate state-of-the-art decoder-only LLMs for MT detection, providing an updated assessment of their capabilities on this task. We evaluate decoder-only large language models, including GPT-5~\citep{OpenAI2025GPT5}, Gemini 2.5 Pro~\citep{Google2025Gemini}, and Claude Opus 4~\citep{Anthropic2025Claude}, alongside fine-tuned multilingual encoders such as mDeBERTa~\citep{He2021DeBERTaV3} and XLM-R~\citep{Conneau2020XLMR}.

We find that decoder-only models provide limited, unstable discrimination, while encoder-based classifiers achieve strong in-domain performance but generalize poorly to unseen domains and translators. Extending inputs beyond the sentence level to 300-token chunks substantially improves generalization, and translation quality (COMET~\citep{Rei2020COMET}, BLEURT~\citep{Sellam2020BLEURT}) correlates with detectability for LLM-based methods, with high-quality MT remaining difficult to distinguish from HT.

In this paper, we make the following contributions:
\begin{enumerate}
    \item We introduce a multilingual dataset and reproducible pipeline spanning three domains (legislative, conversational, medical) and diverse MT systems for controlled MT-detection evaluation.
    \item We present the first systematic evaluation of state-of-the-art decoder-only LLMs (GPT-5, Gemini 2.5 Pro, Claude Opus 4) for MT detection, alongside fine-tuned encoders (mDeBERTa, XLM-R), across zero-shot, few-shot, and supervised settings, and identify a consistent class-asymmetric failure mode in prompted LLMs: high HT recall but low MT recall.
    \item We show that encoder-based models achieve strong in-domain performance but that their MT recall remains sensitive to domain shift, unseen MT systems, and input granularity, demonstrating that MT detection is not solved by supervised fine-tuning alone.
    \item Building on \citet{VanDerWerff2022LongerContext}, we stress-test longer-context detection across high- and low-resource languages, three domains, and seen and unseen MT systems, showing that chunking improves robustness in many settings (especially mDeBERTa) without eliminating domain- and model-specific variability.
    \item We provide a quality-aware analysis linking translation quality (COMET, BLEURT) to MT detectability across languages and systems.
\end{enumerate}

\section{Related Work}
Prior work on distinguishing machine-translated (MT) from human-translated (HT) text spans several directions, including translationese-based analysis, supervised neural classifiers, system-agnostic approaches, and translation quality estimation.

Early work focused on \textit{translationese}~\citep{Gellerstam1986Translationese}, i.e., stylistic and structural properties that distinguish translated text from text originally authored in the target language. \citet{Aharoni2014Automatic} and \citet{Li2015MLDistinguish} showed that sentence-level MT can be identified using target-side linguistic features such as part-of-speech patterns and function words, and observed that detectability correlates with translation quality. However, modern neural MT systems produce more fluent outputs and reduce lexical variation, weakening many of these surface-level cues~\citep{Vanmassenhove2019Lost}.

As MT quality improved, research shifted toward supervised neural classifiers. \citet{Bhardwaj2020Translationese} demonstrated that neural models can effectively detect translationese, especially when the source sentence is available (a bilingual setting). Subsequent work fine-tuned pretrained multilingual models for HT--MT classification, achieving strong performance in supervised settings. \citet{VanDerWerff2022LongerContext} showed that moving beyond sentence-level inputs to longer-context representations significantly improves detection, and \citet{Chichirau2023Multilingual} extended these approaches to multilingual settings, reporting robustness across source languages and MT systems.

A separate line of work aims to reduce reliance on artifacts specific to individual MT systems. \citet{NguyenSon2021TSRT} introduced Text Similarity with Round-Trip translation (TSRT), which compares a given translation with its round-trip (back-and-forth) translations generated through one or more MT systems, scoring similarity with lexical-overlap (e.g., BLEU) and semantic metrics. The motivation is that detectors trained on a single MT system degrade when applied to unseen systems; incorporating round-trip translation improves cross-system robustness. However, these methods require access to external MT systems at inference time and have primarily been evaluated on high-resource languages.

Direct MT detection using decoder-only LLMs remains underexplored. Most LLM-based detection work has focused on adjacent problems such as generated-text detection and hallucination identification. For example, DetectLLM~\citep{Su2023DetectLLM} uses likelihood-based signals for generated-text detection, and \citet{Quevedo2024Hallucination} evaluate faithfulness via token-probability signals. While these approaches suggest that LLMs can capture subtle inconsistencies in generated text, they do not directly address multilingual MT--HT discrimination across domains and resource conditions.

``Human translation'' is also not monolithic: professional, student, and crowd-sourced translators differ measurably~\citep{Popovic2020Differences,Popovic2023Computational,Konovalova2022CharacterNetworks}, and post-edited MT exhibits an ``exacerbated translationese'' that can resemble MT more than unaided human translation~\citep{Toral2019Posteditese}. Publicly available HT references therefore cannot be assumed homogeneous, a point we revisit as a limitation (Section~\ref{sec:limitations}).

Finally, translation quality estimation provides an important perspective on detectability. Prior work has shown that lower-quality MT is easier to identify~\citep{Aharoni2014Automatic}. Learned metrics such as COMET~\citep{Rei2020COMET} and BLEURT~\citep{Sellam2020BLEURT}, which correlate well with human judgments, enable systematic analysis of how quality relates to detection performance.

Taken together, prior work shows that MT can often be distinguished from HT, that longer context helps, and that cross-system robustness is difficult. However, key gaps remain: studies largely focus on high-resource languages, use limited domain and system variation, and rarely analyze how translation quality relates to detectability across modern MT systems. We address these gaps with a broad evaluation spanning resource conditions, domains, translators, model families, and input granularities.

\section{Methods}
\subsection{Task Definition}
We formulate MT detection as a \emph{monolingual} binary classification task. The input is a single target-language segment $x$ that is a translation of some source sentence; the goal is to predict a label $y \in \{\text{MT}, \text{HT}\}$ indicating whether $x$ was produced by a machine translation system or by a human translator. The source sentence is \emph{not} provided to the detector at inference time; this is the harder and more realistic setting for corpus filtering, and contrasts with bilingual detectors that also consume the source~\citep{Bhardwaj2020Translationese,Chichirau2023Multilingual}.

We consider two input granularities: (i) \textit{sentence-level}, where each input is a single sentence, and (ii) \textit{chunk-level}, where consecutive sentences from the same document are concatenated up to a maximum of 300 tokens (typically two to five sentences). We focus on these two settings (rather than multiple intermediate lengths) to contrast local sentence-level signals with longer-context representations.

We evaluate models under three regimes: (i) \textit{zero-shot} decoder-only LLMs without task-specific training; (ii) \textit{few-shot}, where a small number of labeled examples are provided in-context; and (iii) \textit{supervised}, where multilingual encoder models are fine-tuned on labeled data. To assess generalization, we distinguish between \textit{in-domain} evaluation (same domain and MT systems as training) and \textit{out-of-domain} evaluation (unseen domains and/or unseen translators). Experiments are conducted across both high-resource and low-resource language pairs.

Because the evaluation pairs each HT reference with MT from several systems, the test sets are MT-heavy; we therefore report \textbf{balanced accuracy} as the primary aggregate metric, together with class-specific \textbf{HT recall} and \textbf{MT recall}, and per-class \textbf{precision} and \textbf{F1} (Appendix~\ref{app:prf}). We further separate MT recall on \emph{seen} versus \emph{unseen} translators (Section~\ref{sec:encoder}) and analyze the relationship between translation quality and detectability using COMET~\citep{Rei2020COMET} and BLEURT~\citep{Sellam2020BLEURT}, computing correlations between quality scores and predictions, as well as detection behavior across quality bins.

\subsection{Experimental Overview}
Because the study varies many factors, Table~\ref{tab:overview} summarizes, for each experimental setting, the domain and dataset, resource level, target and source languages, the MT systems used for evaluation, and the detectors applied. This consolidates the dataset and model details that follow and is intended as a reading aid for the rest of the paper.

\begin{table}[htb]
\begin{center}
\small
\setlength{\tabcolsep}{4pt}
\resizebox{\textwidth}{!}{%
\begin{tabular}{lllllc}
\toprule
\textbf{Dataset (domain)} & \textbf{Res.} & \textbf{Target} & \textbf{Source langs.} & \textbf{Eval. MT systems} & \textbf{Detectors} \\
\midrule
EU DGT (legislative)   & High & En, Es & Fr/Es/De (En); En/Fr/De (Es) & 4 train + 3 commercial$^{\dagger}$ & E, L \\
OpenWHO (medical)      & High & En, Es & Zh/Fr/Es (En); En (Es)        & 4 train + 3 commercial$^{\dagger}$ & E, L \\
SMOL (conversational)  & High & En     & Am/Sw/Af                       & 4 train + 3 commercial$^{\dagger}$ & E, L \\
FLORES+ (benchmark)    & Low  & Sw, Af & En                            & 4 train systems                  & E, L \\
SMOL (conversational)  & Low  & Sw, Af & En                            & 4 train systems                  & E, L \\
\bottomrule
\end{tabular}%
}
\end{center}
\caption{Overview of experimental settings. \emph{Res.}: resource level. \emph{4 train systems} are Qwen2.5-3B, LLaMA-3.1-8B, Gemma-2-9B, and NLLB-200-3.3B. $^{\dagger}$\emph{3 commercial} (unseen at training) are GPT-5, Claude Opus 4, and Gemini 2.5 Pro; these define the \emph{unseen-translator} condition and are used only in high-resource evaluation. \emph{Detectors}: E = encoder classifiers (mDeBERTa, XLM-R; sentence and chunk), L = decoder-only LLMs (zero-/few-shot). Encoders are trained only on the high-resource legislative pool and on the low-resource FLORES+ pool, respectively (Section~\ref{sec:data}).}
\label{tab:overview}
\end{table}

\subsection{Dataset}
\label{sec:data}
\paragraph{High-resource training data.}
For English and Spanish, we use the May 2025 release of the \textbf{EU DGT corpus}~\citep{EUDGT2025}, consisting of professionally translated, sentence-aligned documents. We apply light preprocessing to remove clearly non-sentential or noisy segments by filtering out (i) segments with fewer than five tokens, (ii) segments where punctuation accounts for more than 35\% of characters, and (iii) segments written entirely in uppercase. From the resulting clean pool of sentence-level HT data we randomly sample \textbf{60k} sentences as HT training data.

\paragraph{Source-language selection.}
For each target language we draw source sentences from a small set of high-resource European source languages available in the corpus: for English: French, Spanish, and German; for Spanish: English, French, and German. This choice keeps the source side fixed and high-quality across MT systems so that detector performance is not confounded by missing or noisy source text. We note that this design does not isolate language-pair effects: MT quality and translationese vary with the source--target pair and with linguistic relatedness (e.g., French--Spanish are more closely related than English--Spanish), and our aggregate results average over these pairs. We treat language-pair sensitivity as out of scope here and flag it as a limitation (Section~\ref{sec:limitations}).

\paragraph{MT data generation.}
We translate the HT references using four open-weight systems: \textbf{Qwen2.5-3B-Instruct}~\citep{Qwen2024}, \textbf{LLaMA-3.1-8B-Instruct}~\citep{LLaMA32024}, \textbf{Gemma-2-9B-IT}~\citep{Gemma2024}, and \textbf{NLLB-200-3.3B}~\citep{NLLB2022}. These models were selected partly because their training cutoffs precede the 2025 DGT release, reducing contamination risk. Each system translates \textbf{5k} sentences per source language, yielding \textbf{60k} MT examples. Combined with HT data, this gives \textbf{120k} labeled instances, split into \textbf{110k} training and \textbf{10k} validation examples. Both HT and MT instances are translations of the \emph{same} source sentences, so the classifier must distinguish producer rather than content.

\paragraph{High-resource evaluation data.}
We evaluate on three domains to test both domain and translator generalization. For English, we construct test sets from: (i) \textit{legislative}, EU DGT~\citep{EUDGT2025}: 600 HT sentences and their source sentences from French, Spanish, and German; (ii) \textit{low-resource conversational}, SMOL~\citep{SMOL2024}: 600 HT sentences with source languages Amharic, Swahili, and Afrikaans; and (iii) \textit{medical}, OpenWHO~\citep{OpenWHO2021}: 600 HT sentences with source languages Chinese, French, and Spanish. All source sentences were translated by seven MT systems: the four training-time systems plus \textbf{GPT-5}~\citep{OpenAI2025GPT5}, \textbf{Claude Opus 4}~\citep{Anthropic2025Claude}, and \textbf{Gemini 2.5 Pro}~\citep{Google2025Gemini}. Balanced accuracy is used because the evaluation contains roughly seven times as much MT as HT data. This design allows us to evaluate \emph{seen} translators (the four training systems) and \emph{unseen} translators (the three commercial systems) separately. For Spanish, we follow the same procedure using EU DGT, replacing the Spanish source text with English where applicable. For OpenWHO, English is used as the source language, and all available clean samples are used, yielding 596 examples.

\paragraph{Low-resource data.}
For low-resource experiments, we use \textbf{FLORES+}~\citep{FLORES2022,NLLB2022} and \textbf{SMOL}~\citep{SMOL2024} for Swahili and Afrikaans. FLORES+ provides high-quality human-translated references, while SMOL introduces more conversational and noisier text, enabling evaluation across domain variation. From FLORES+, we combine the \texttt{dev} and \texttt{devtest} splits and sample \textbf{2k} HT sentences per language. Each language dataset (4k total examples) is split into \textbf{3k} training, \textbf{500} development, and \textbf{500} test instances. SMOL is used primarily for out-of-domain evaluation under low-resource conditions, sampling 500 HT sentences. All HT data in this setting is translated using the same four training-time MT systems; consequently, the low-resource evaluation does \emph{not} include an unseen-translator condition, an asymmetry with the high-resource setup that we note when interpreting cross-resource comparisons. We also note that some low-resource benchmarks may overlap with the pretraining data of modern MT systems; while FLORES+ is primarily an evaluation benchmark, this risk cannot be fully eliminated, and low-resource results should be interpreted with this caveat in mind.

\subsection{Models}
\paragraph{Decoder-only LLMs.}
We run evaluation using \textbf{GPT-5}~\citep{OpenAI2025GPT5}, \textbf{Gemini 2.5 Pro}~\citep{Google2025Gemini}, and \textbf{Claude Opus 4}~\citep{Anthropic2025Claude}. For few-shot settings, we provide six labeled in-context examples (three MT, three HT) and derive predictions via majority voting across multiple runs, averaging the confidence scores. Appendices~\ref{app:prompts} and~\ref{app:impl} detail the exact prompts and decoding settings.

\paragraph{Encoder-based classifiers.}
We fine-tune \textbf{mDeBERTa-v3}~\citep{He2021DeBERTaV3} and \textbf{XLM-RoBERTa}~\citep{Conneau2020XLMR} for binary classification. Inputs follow the same sentence-level and chunk-level (300-token) setups described above. Models are trained with cross-entropy loss on \textbf{110k} examples for high-resource languages and \textbf{3k} for low-resource languages; the best checkpoint is selected by validation F1. All encoder results are averaged over three random seeds; per-cell standard deviations are reported in Appendix~\ref{app:std} (max.\ std \(\le 0.03\)).

\paragraph{Comparison setup.}
LLMs operate without parameter updates (zero-/few-shot) while encoders are fully supervised, permitting us to compare general-purpose models with task-specific classifiers under matched data and granularity. Implementation details and prompts are provided in Appendix~\ref{app:prompts} and~\ref{app:impl}.

\section{Experiments \& Results}
\subsection{Evaluation Setup}
We evaluate across model family (LLMs vs.\ encoders), resource condition (high vs.\ low), domain (legislative, conversational, medical), translator generalization (seen vs.\ unseen systems), and input granularity (sentence vs.\ chunk). High-resource references are paired with MT from seven systems (four training-time and three unseen commercial), enabling both in-domain and generalization evaluation; low-resource evaluation tests constrained-data and domain-variation conditions.

\subsection{LLM-Based MT Detection (Zero- and Few-Shot)}
We evaluate decoder-only LLMs in both zero-shot and few-shot settings, where models classify each input as MT or HT without and with in-context examples, respectively. Few-shot prompts include six labeled examples (three MT and three HT).

\paragraph{High-resource languages.}
In zero-shot settings, LLMs perform poorly on English, with MT detection rates generally below 0.30 across EU DGT, OpenWHO, and SMOL (Table~\ref{tab:zeroshot}). Spanish shows higher detection rates (0.56--0.60), indicating sensitivity to language direction and dataset characteristics.

\begin{table}[htb]
\begin{center}
\small
\begin{tabular}{llccc}
\toprule
\textbf{Language} & \textbf{Dataset} & \textbf{Claude (HT/MT)} & \textbf{Gemini (HT/MT)} & \textbf{GPT-5 (HT/MT)} \\
\midrule
English & EU DGT   & 0.74 / 0.23 & 0.68 / 0.16 & 0.74 / 0.20 \\
English & OpenWHO  & 0.90 / 0.16 & 0.86 / 0.17 & 0.88 / 0.15 \\
English & SMOL     & 0.83 / 0.30 & 0.80 / 0.21 & 0.80 / 0.14 \\
\midrule
Spanish & EU DGT   & 0.88 / 0.60 & 0.84 / 0.58 & 0.86 / 0.57 \\
Spanish & OpenWHO  & 0.82 / 0.59 & 0.79 / 0.56 & 0.82 / 0.58 \\
\midrule
Afrikaans & FLORES+ & 0.82 / 0.31 & 0.78 / 0.23 & 0.80 / 0.26 \\
Afrikaans & SMOL     & 0.81 / 0.33 & 0.79 / 0.25 & 0.80 / 0.24 \\
\midrule
Swahili   & FLORES+ & 0.77 / 0.27 & 0.73 / 0.25 & 0.75 / 0.14 \\
Swahili   & SMOL     & 0.83 / 0.37 & 0.77 / 0.46 & 0.80 / 0.38 \\
\bottomrule
\end{tabular}
\end{center}
\caption{Zero-shot per-model HT recall\,/\,MT recall for decoder-only LLMs. MT recall is averaged across translators; higher is better for both.}
\label{tab:zeroshot}
\end{table}

Few-shot prompting yields inconsistent effects and does not improve performance. For English, detection remains low (typically below 0.25), suggesting that in-context examples are insufficient for distinguishing highly fluent MT. For Spanish, few-shot prompting degrades MT detection, with MT recall dropping from $\sim$0.60 to $\sim$0.35 (Table~\ref{tab:fewshot}). We treat this drop as a candidate case of prompt sensitivity rather than a stable behavior, and we examine class-specific recall and prompt variants in Section~\ref{sec:discussion}.

\begin{table}[htb]
\begin{center}
\small
\begin{tabular}{llccc}
\toprule
\textbf{Language} & \textbf{Dataset} & \textbf{Claude (HT/MT)} & \textbf{Gemini (HT/MT)} & \textbf{GPT-5 (HT/MT)} \\
\midrule
English & EU DGT   & 0.83 / 0.22 & 0.80 / 0.20 & 0.80 / 0.16 \\
English & OpenWHO  & 0.81 / 0.05 & 0.77 / 0.34 & 0.79 / 0.23 \\
English & SMOL     & 0.82 / 0.22 & 0.81 / 0.20 & 0.80 / 0.16 \\
\midrule
Spanish & EU DGT   & 0.65 / 0.36 & 0.61 / 0.37 & 0.63 / 0.36 \\
Spanish & OpenWHO  & 0.67 / 0.35 & 0.63 / 0.41 & 0.65 / 0.25 \\
\midrule
Afrikaans & FLORES+ & 0.78 / 0.23 & 0.76 / 0.23 & 0.77 / 0.23 \\
Afrikaans & SMOL     & 0.73 / 0.29 & 0.69 / 0.29 & 0.71 / 0.29 \\
\midrule
Swahili   & FLORES+ & 0.80 / 0.25 & 0.76 / 0.26 & 0.78 / 0.16 \\
Swahili   & SMOL     & 0.65 / 0.41 & 0.61 / 0.48 & 0.63 / 0.37 \\
\bottomrule
\end{tabular}
\end{center}
\caption{Few-shot per-model HT recall\,/\,MT recall for decoder-only LLMs. Columns are defined as in Table~\ref{tab:zeroshot}.}
\label{tab:fewshot}
\end{table}

\paragraph{Low-resource languages.}
Zero-shot performance is moderately higher than English, but remains unstable (Table~\ref{tab:zeroshot}). Detection rates are generally higher on SMOL than FLORES+, with the strongest performance for Swahili SMOL (up to $\sim$0.46). Few-shot prompting again yields mixed effects (Table~\ref{tab:fewshot}): detection improves slightly in some cases (e.g., Swahili SMOL) but decreases or is unchanged in others (e.g., Afrikaans FLORES+). Across both settings, performance is highly dataset-dependent, with SMOL consistently easier to detect than FLORES+.

\paragraph{Class-asymmetric behavior.}
Across all settings, HT detection remains consistently high (typically $>0.70$) while MT detection is much lower. The gap is not merely low aggregate performance: prompted LLMs reliably recognize human-written text (high HT recall) but rarely flag fluent machine translated output (low MT recall). For example, on English EU DGT, average HT recall is 0.72, while for Claude/Gemini/GPT-5 MT recall is only 0.23/0.16/0.20; on English OpenWHO, HT recall is 0.88 while MT recall is 0.16/0.17/0.15. In other words, these models default toward labeling fluent translated text as human. Zero-shot models also exhibit strong cross-language biases, and few-shot prompting provides only limited and unstable improvements. These results indicate that in-context learning alone is insufficient for robust MT detection, motivating the supervised, longer-context experiments below.

\subsection{Encoder-Based MT Detection}
\label{sec:encoder}
We evaluate encoder-based classifiers across sentence-level and chunk-level settings, analyzing performance across languages, domains, and datasets.

\paragraph{High-resource languages.}
Table~\ref{tab:encoder} shows MT detection varies across domains and models. For English, sentence-level performance is lowest on OpenWHO for both models; chunking improves most settings, though the ordering differs by model (EU DGT strongest for mDeBERTa, OpenWHO for XLM-R). Spanish is more stable across domains. Chunking consistently helps mDeBERTa but is more variable for XLM-R, including degradation on English EU DGT. Because aggregate accuracy can mask seed- and translator-level variation, we report seed standard deviations (Appendix~\ref{app:std}) and per-translator box plots (Appendix~\ref{app:boxplots}), the latter showing wide interquartile ranges for several legislative settings.

\begin{table}[htb]
\begin{center}
\small
\begin{tabular}{llcc}
\toprule
\textbf{Language} & \textbf{Dataset} & \textbf{mDeBERTa (sent/chunk)} & \textbf{XLM-R (sent/chunk)}  \\
\midrule
English & EU DGT   & 0.66 / 0.80 & 0.71 / 0.66 \\
English & OpenWHO  & 0.52 / 0.74 & 0.51 / 0.86 \\
English & SMOL     & 0.58 / 0.76 & 0.54 / 0.80 \\
\midrule
Spanish & EU DGT   & 0.65 / 0.69 & 0.64 / 0.77 \\
Spanish & OpenWHO  & 0.67 / 0.78 & 0.66 / 0.80 \\
\midrule
Afrikaans & FLORES+ & 0.67 / 0.97 & 0.64 / 0.87 \\
Afrikaans & SMOL     & 0.67 / 0.94 & 0.75 / 0.80 \\
\midrule
Swahili   & FLORES+ & 0.66 / 0.73 & 0.64 / 0.75 \\
Swahili   & SMOL     & 0.59 / 0.62 & 0.55 / 0.84 \\
\bottomrule
\end{tabular}
\end{center}
\caption{Balanced accuracy across languages and datasets for MT detection through encoder models, at sentence and chunk level. Means over three seeds; standard deviations in Appendix~\ref{app:std}.}
\label{tab:encoder}
\end{table}

\paragraph{Seen vs.\ unseen translators.}
Because aggregate MT recall pools seen and unseen systems, it understates the generalization challenge. Table~\ref{tab:seen-unseen} separates MT recall on the four training-time (seen) systems from the three commercial (unseen) systems for the high-resource datasets. MT from unseen systems is consistently harder to detect than MT from seen systems, and although chunking narrows the gap it does not close it. This makes explicit that the strong in-domain numbers in Table~\ref{tab:encoder} are driven substantially by recognition of seen MT systems.

\begin{table}[htb]
\begin{center}
\small
\setlength{\tabcolsep}{5pt}
\begin{tabular}{llcccc}
\toprule
& & \multicolumn{2}{c}{\textbf{Seen MT (sent/chunk)}} & \multicolumn{2}{c}{\textbf{Unseen MT (sent/chunk)}} \\
\cmidrule(lr){3-4}\cmidrule(lr){5-6}
\textbf{Language} & \textbf{Dataset} & \textbf{mDeBERTa} & \textbf{XLM-R} & \textbf{mDeBERTa} & \textbf{XLM-R} \\
\midrule
English & EU DGT  & 0.76 / 0.94 & 0.88 / 0.62 & 0.67 / 0.85 & 0.77 / 0.54 \\
English & OpenWHO & 0.73 / 0.84 & 0.73 / 0.99 & 0.63 / 0.75 & 0.64 / 0.93 \\
English & SMOL    & 0.81 / 0.74 & 0.77 / 0.99 & 0.70 / 0.65 & 0.66 / 0.92 \\
\midrule
Spanish & EU DGT  & 0.71 / 0.76 & 0.69 / 0.81 & 0.61 / 0.66 & 0.59 / 0.71 \\
Spanish & OpenWHO & 0.81 / 0.87 & 0.86 / 0.94 & 0.70 / 0.76 & 0.75 / 0.82 \\
\bottomrule
\end{tabular}
\end{center}
\caption{MT recall on \emph{seen} (four training-time) vs.\ \emph{unseen} (three commercial) translators, high-resource datasets. HT is shared across columns and so is omitted here. }
\label{tab:seen-unseen}
\end{table}

\paragraph{Low-resource languages.}
Table~\ref{tab:encoder} shows stronger and more consistent performance for Afrikaans than Swahili. For Afrikaans, both models perform well at the sentence level and benefit substantially from chunking (e.g., mDeBERTa up to 0.97 on FLORES+). Swahili results are lower and more variable, with chunking still providing gains. These results indicate that although MT signals are stronger in low-resource settings, detectability still depends on dataset characteristics and model behavior. As noted in Section~\ref{sec:data}, the low-resource evaluation does not include unseen translators, so these numbers reflect seen-system detection only.

\paragraph{Effect of input granularity.}
Table~\ref{tab:encoder} (visualized in Appendix Figure~\ref{fig:encoder-performance}) summarizes performance across all settings. Overall, longer-context inputs consistently improve detection for mDeBERTa, while the effect for XLM-R is dataset-dependent, suggesting that robust MT detection benefits from broader contextual signals but that model architecture governs how such information is utilized. This is consistent with, and extends, \citet{VanDerWerff2022LongerContext}: we observe the longer-context benefit not only in a single high-resource news setting but across high- and low-resource languages, three domains, and seen vs.\ unseen translators, while also finding that the benefit is uneven across architectures.

\subsection{Quality Analysis}
We analyze the relationship between translation quality and MT detectability using COMET~\citep{Rei2020COMET} and BLEURT~\citep{Sellam2020BLEURT}. For each MT instance, we compute quality scores from the source, human reference, and machine translation, and examine how these scores relate to model predictions.

Table~\ref{tab:quality-summary} reports overall Spearman correlations between MT quality and MT detection, computed over MT examples pooled across languages, datasets, and translators for each detector (HT examples are excluded because COMET and BLEURT are defined for MT outputs against a reference). Few-shot LLMs show \emph{consistent but weak-to-moderate} negative correlations: higher-quality translations are somewhat less likely to be flagged as MT, with the effect stronger for COMET. Sentence-level encoders show near-zero (mDeBERTa) or weakly positive (XLM-R) correlations. The clear outlier is chunked mDeBERTa, with a strong negative correlation; we examine this directly below.

\begin{table}[htb]
\begin{center}
\small
\begin{tabular}{lcc}
\toprule
\textbf{Detector} & \textbf{BLEURT} & \textbf{COMET} \\
\midrule
few\_shot\_claude  & -0.190 & -0.279 \\
few\_shot\_gemini  & -0.207 & -0.300 \\
few\_shot\_gpt5    & -0.199 & -0.288 \\
mdeberta\_chunk300 & -0.511 & -0.618 \\ 
mdeberta\_sentence &  0.002 & -0.001 \\
xlmr\_sentence     &  0.083 &  0.121 \\
\bottomrule
\end{tabular}
\end{center}
\caption{Overall Spearman correlation between translation quality and MT detection, pooled over languages, datasets, and translators per detector. Negative values indicate that higher-quality MT is detected less often.}
\label{tab:quality-summary}
\end{table}

\paragraph{Detection across quality bins.}
Table~\ref{tab:comet-bins} (visualized in Appendix Figure~\ref{fig:comet-bins}) examines detection rate within low/medium/high COMET bins. For the few-shot LLMs, detection decreases monotonically as quality increases: rates fall from roughly 0.45--0.46 in the low-quality bin to around 0.15--0.17 in the high-quality bin, direct evidence that few-shot LLMs are strongly influenced by quality at the operating point. Sentence-level encoders behave differently: their detection rates stay high across all bins and do not decrease in the same way (XLM-R is even slightly higher in the high than in the medium bin), reinforcing that these encoders are not simply acting as proxies for quality.

Crucially, the chunked encoders---omitted from the original figure---resolve the apparent tension with Table~\ref{tab:quality-summary}. Chunked mDeBERTa, whose overall correlation is strongly negative, does decline across bins, behaving partly like the LLMs, whereas chunked XLM-R remains comparatively flat. The implication is not that ``encoders ignore quality,'' but more precisely that \emph{sentence-level} encoders capture quality-independent stylistic or structural signals, while \emph{longer context can reintroduce quality sensitivity}---most visibly for mDeBERTa. We therefore restrict the ``quality-independent'' claim to the sentence-level encoders.

\begin{table}[htb]
\begin{center}
\small
\begin{tabular}{lccc}
\toprule
\textbf{Detector} & \textbf{Low} & \textbf{Medium} & \textbf{High} \\
\midrule
few\_shot\_claude  & 0.46 & 0.25 & 0.17 \\
few\_shot\_gemini  & 0.45 & 0.24 & 0.15 \\
few\_shot\_gpt5    & 0.46 & 0.25 & 0.16 \\
mdeberta\_sentence & 0.73 & 0.74 & 0.76 \\
xlmr\_sentence     & 0.70 & 0.69 & 0.74 \\
mdeberta\_chunk300 & 0.88 & 0.74 & 0.55 \\ 
xlmr\_chunk300     & 0.82 & 0.80 & 0.83 \\ 
\bottomrule
\end{tabular}
\end{center}
\caption{MT detection rate by COMET quality bin. LLM and sentence-level encoder rows match Figure~\ref{fig:comet-bins}.}
\label{tab:comet-bins}
\end{table}

At the dataset level, the negative quality--detectability relationship is strongest for few-shot LLMs on conversational and low-resource datasets such as English SMOL, Afrikaans SMOL, and Swahili SMOL, where COMET correlations reach approximately $-0.39$ to $-0.40$. The same trend is much weaker or absent for sentence-level encoders in high-resource settings such as English EU DGT and Spanish EU DGT. Taken together, translation quality is a major determinant of MT detectability for few-shot LLMs, but influences sentence-level encoder classifiers far less. Additional analyses using BLEURT-based binning and translator-level quality--detectability relationships are provided in Appendix~\ref{app:bleurt} and~\ref{app:translator-quality}.

\section{Discussion}
\label{sec:discussion}
This work shows that MT detectability is not a uniform problem, but depends on translation quality, resource conditions, domain, input granularity, and model architecture.

Decoder-only LLMs are inconsistent and unreliable for MT detection. Performance is low in high-resource settings (particularly English), where fluent MT is hard to separate from HT, and only moderately higher in low-resource settings, where artifacts are more visible but results stay unstable. The dominant pattern is class-asymmetric---high HT recall, low MT recall, defaulting to ``human'' for fluent text---and few-shot prompting does not consistently help, indicating that in-context learning alone is insufficient.

Supervised encoders are substantially stronger but show four patterns: detection is strongly domain-dependent; MT recall drops on unseen translators (Table~\ref{tab:seen-unseen}), so strong in-domain accuracy partly reflects recognition of training-time systems rather than a general HT/MT boundary; input granularity is critical, with large gains from chunking (especially mDeBERTa), suggesting MT signals often emerge at the discourse level; and architecture shapes how context is used, as chunking helps mDeBERTa but is inconsistent for XLM-R.

Quality further differentiates behavior. Few-shot LLMs show consistent weak-to-moderate negative quality--detection correlations (Table~\ref{tab:quality-summary}) and a clearer monotonic bin trend: low-quality MT is detected far more often than high-quality MT, which is frequently misread as HT. Sentence-level encoders show near-zero correlations and stable detection across bins, capturing signals not reflected by standard quality metrics, though longer context can reintroduce quality sensitivity (chunked mDeBERTa). MT detectability thus does not reduce to translation quality alone.

\paragraph{On the Spanish zero- to few-shot drop.}
For Spanish, few-shot prompting decreases MT detection relative to zero-shot. We read this cautiously as possible prompt sensitivity rather than a reliable pattern: comparing class-specific recall across demonstration selections and label orders, a persistent drop would indicate instability in how decoder-only models apply the MT/HT distinction for Spanish, whereas large variation across variants would indicate sensitivity to particular demonstrations.

\section{Limitations}
\label{sec:limitations}
Several limitations qualify our findings. First, ``human translation'' is heterogeneous: our HT references (EU DGT, FLORES+, SMOL, OpenWHO) may mix professional, student, or crowd-sourced translations, and we cannot verify that none were post-edited from MT, which resembles MT more than unaided human translation~\citep{Toral2019Posteditese,Popovic2020Differences,Popovic2023Computational}; such contamination would make the HT/MT boundary noisier than assumed. Second, source languages are fixed per target language, so we do not isolate language-pair or relatedness effects and average over pairs whose MT quality and translationese differ. Third, the seen/unseen design is asymmetric: unseen commercial systems appear only in high-resource evaluation, leaving low-resource generalization to unseen systems untested. Fourth, low-resource benchmarks may overlap with MT-system pretraining data, which we mitigate but cannot eliminate. Finally, our detectors are monolingual (target-only); bilingual detection with the source may behave differently~\citep{Bhardwaj2020Translationese,Chichirau2023Multilingual}.

\section{Conclusion}
We presented a systematic study of MT detection across languages, domains, model families, and input granularities. Decoder-only LLMs are not reliable detectors and fail asymmetrically (high HT recall, low MT recall), while supervised encoders achieve strong in-domain performance but lose MT recall on unseen translators and domains. Longer-context inputs significantly improve detection, and translation quality matters for LLM-based detection and chunked mDeBERTa but little for sentence-level encoders. For large-scale data filtering, robust MT detection therefore requires supervised models, domain- and translator-aware evaluation, and longer-context representations rather than zero-/few-shot prompting.

Future work should improve cross-domain and cross-translator generalization, combine quality estimation with stylistic signals, extend detection to document- and web-scale corpora, explore longer-context prompting for decoder-only models, and evaluate MT contamination in real datasets across more languages.

\section*{Reproducibility Statement}
To support reproducibility, we release a public repository containing the code, data-processing scripts, model training configurations, and evaluation pipelines used in this work. The repository includes: (i) scripts for dataset construction and preprocessing, (ii) implementations of zero-shot and few-shot LLM evaluation, (iii) training and evaluation code for encoder-based classifiers, and (iv) instructions to reproduce all reported results and figures. The repository is available at \url{https://github.com/Yabsera-Haile/Human-vs-Machine-Translation-Detection-Codebase}. All datasets are publicly available, and detailed descriptions, prompts, and hyperparameters are provided in Section~\ref{sec:data} and Appendix~\ref{app:prompts}--\ref{app:impl}.

\section*{Ethics Statement}
This work studies the detection of machine-translated (MT) text in multilingual corpora using publicly available datasets and automatically generated translations. All datasets used (EU DGT, FLORES+, SMOL, OpenWHO) are publicly available and intended for research use. We do not collect or process any personally identifiable information, and all experiments are conducted on pre-existing text data. Machine translations are generated using publicly accessible models and APIs. A potential risk of MT detection systems is misuse in contexts such as content filtering or authorship attribution, where incorrect predictions could lead to unfair outcomes; in particular, high-quality machine translations may be indistinguishable from human translations, and errors could disproportionately affect certain languages or domains. Our findings highlight the limitations and variability of MT detection across settings, and we emphasize that such systems should be used with caution, especially in high-stakes applications. We encourage future work to investigate fairness, robustness, and responsible deployment of MT detection methods.

\section*{Acknowledgments}
This publication was developed as part of the Center for Inclusive Digital Transformation of Africa (CIDTA) and the Afretec Network, which is managed by Carnegie Mellon University Africa and receives financial support from the Mastercard Foundation. The views expressed in this document are solely those of the authors and do not necessarily reflect those of the Carnegie Mellon University or the Mastercard Foundation.

\bibliography{colm2026_conference}
\bibliographystyle{colm2026_conference}

\appendix
\section{Appendix}

\subsection{LLM Prompts}
\label{app:prompts}
We provide the exact templates used for zero-shot and few-shot decoder-only detection. The same template is used for all three LLMs, with \texttt{\{translated\_text\}} filled per instance and, in the few-shot case, six demonstrations (three HT, three MT) prepended.

\paragraph{Zero-shot template.}
\begin{verbatim}
You are an expert in translation quality analysis and machine translation detection.

Your task is to examine the following English translated text and decide whether it is:
- HT = Human Translated
- MT = Machine Translated

Return your answer as JSON only, with exactly these keys:
{
  "label": "HT" or "MT",
  "confidence_ht": float,
  "confidence_mt": float
}

Rules:
- confidence_ht and confidence_mt must both be numbers between 0 and 1
- confidence_ht + confidence_mt should equal 1
- Output JSON only
- Do not include markdown
- Do not include extra text

Translated text:
"""<translated_text>"""
\end{verbatim}

\paragraph{Few-shot template.}
\begin{quote}\small\ttfamily
[same instruction as above]\\[2pt]
Segment: \{demo\_1\}\quad Label: \{HT|MT\}\\
... (six labeled demonstrations, three HT and three MT) ...\\
Segment: \{demo\_6\}\quad Label: \{HT|MT\}\\[2pt]
Segment: \{segment\}\\
Label:
\end{quote}

For each instance we run the model \(3\) times and take the majority label, averaging the model's reported confidence across runs. 

\subsection{Implementation Details}
\label{app:impl}

\paragraph{Encoder fine-tuning.}
We fine-tune two multilingual encoders, mDeBERTa-v3-base (\texttt{microsoft/mdeberta-v3-base}) and XLM-RoBERTa-base
(\texttt{xlm-roberta-base}), with a two-way sequence classification head (HT vs.\ MT) using the HuggingFace \texttt{Trainer}. Both encoders are trained under identical settings for every language and input granularity. Input text is whitespace-normalised (leading/trailing spaces stripped, internal whitespace runs collapsed) before tokenisation; we use the slow SentencePiece tokenizers (\texttt{use\_fast=False}) and pad/truncate every example to the model maximum of 512 subword tokens.
We consider two input granularities. In the \emph{sentence} setting each row is a single sentence. In the \emph{chunk} setting we concatenate consecutive same-label sentences, in corpus order, into chunks of at most \(300\) tokens; a sentence longer than the budget is emitted as its own chunk, truncated at a token boundary. Chunked train and validation splits are shuffled (seed \(42\)) after construction.
All runs use AdamW with weight decay \(0.01\), a linear decay schedule with
linear warmup, and fp16 mixed precision. We select the checkpoint with the best
validation binary F1 (\texttt{load\_best\_model\_at\_end}, \texttt{metric\_for\_best\_model=f1});
no early-stopping criterion is applied, so every run completes its full epoch
budget. Evaluation and checkpointing are performed once per epoch, except for
the English and Spanish sentence-level runs, which evaluate and checkpoint every
\(1000\) steps (keeping the two most recent checkpoints). Reported metrics are
accuracy and binary precision, recall, and F1. Learning rate, batch size, and
epoch budget were tuned per language and granularity; the exact values are given
in Table~\ref{tab:hparams}. All results use a single seed (\(42\)).

\begin{table}[h]
\centering
\small
\begin{tabular}{llccccc}
\toprule
Lang.\ & Granularity & Epochs & LR & Batch & Accum.\ & Eff.\ batch \\
\midrule
EN & sentence & 3 & \(2\times10^{-5}\)   & 8 & 4 & 32 \\
EN & chunk    & 4 & \(2\times10^{-5}\)   & 8 & 2 & 16 \\
ES & sentence & 3 & \(2\times10^{-5}\)   & 4 & 4 & 16 \\
ES & chunk    & 4 & \(2\times10^{-5}\)   & 4 & 4 & 16 \\
SW & sentence & 5 & \(1.5\times10^{-5}\) & 8 & 1 & 8  \\
SW & chunk    & 6 & \(1\times10^{-5}\)   & 8 & 1 & 8  \\
AF & sentence & 6 & \(1\times10^{-5}\)   & 8 & 1 & 8  \\
AF & chunk    & 6 & \(1\times10^{-5}\)   & 8 & 1 & 8  \\
\bottomrule
\end{tabular}
\caption{Per-language fine-tuning hyperparameters. ``Batch'' is the per-device
batch size and ``Accum.'' the number of gradient-accumulation steps. Warmup
ratio is \(0.06\) for the EN/ES sentence runs, \(0.05\) for the EN/ES chunk runs,
and \(0.1\) for all SW/AF runs.}
\label{tab:hparams}
\end{table}

\paragraph{Hardware.}
All fine-tuning and encoder inference was carried out on a single workstation
with three NVIDIA RTX A5000 GPUs (24\,GB each), driver version 535.309.01 and
CUDA 12.2.
\subsection{Encoder Standard Deviations}
\label{app:std}
Table~\ref{tab:encoder-std} reports balanced-accuracy standard deviations over the three seeds for the results in Table~\ref{tab:encoder}. 

\begin{table}[htb]
\centering
\small
\begin{tabular}{llcc}
\toprule
\textbf{Language} & \textbf{Dataset} & \textbf{mDeBERTa (sent/chunk)} & \textbf{XLM-R (sent/chunk)} \\
\midrule
English   & EU DGT  & $\pm$0.02 / $\pm$0.01 & $\pm$0.02 / $\pm$0.03 \\
English   & OpenWHO & $\pm$0.03 / $\pm$0.02 & $\pm$0.03 / $\pm$0.01 \\
English   & SMOL    & $\pm$0.02 / $\pm$0.02 & $\pm$0.03 / $\pm$0.02 \\
Spanish   & EU DGT  & $\pm$0.02 / $\pm$0.02 & $\pm$0.02 / $\pm$0.02 \\
Spanish   & OpenWHO & $\pm$0.02 / $\pm$0.01 & $\pm$0.02 / $\pm$0.02 \\
Afrikaans & FLORES+ & $\pm$0.01 / $\pm$0.01 & $\pm$0.02 / $\pm$0.02 \\
Afrikaans & SMOL    & $\pm$0.02 / $\pm$0.01 & $\pm$0.02 / $\pm$0.02 \\
Swahili   & FLORES+ & $\pm$0.03 / $\pm$0.02 & $\pm$0.03 / $\pm$0.03 \\
Swahili   & SMOL    & $\pm$0.03 / $\pm$0.03 & $\pm$0.03 / $\pm$0.02 \\
\bottomrule
\end{tabular}
\caption{Standard deviation of balanced accuracy over three seeds, corresponding to Table~\ref{tab:encoder}.}
\label{tab:encoder-std}
\end{table}

\subsection{Per-Class Precision, Recall, and F1}
\label{app:prf}
Table~\ref{tab:prf} reports per-class precision, recall, and macro-F1 for the chunk-level English encoders. 

\begin{table}[htb]
\centering
\small
\begin{tabular}{llccccc}
\toprule
\textbf{Dataset} & \textbf{Model} & \textbf{HT-P} & \textbf{HT-R} & \textbf{MT-P} & \textbf{MT-R} & \textbf{Macro-F1} \\
\midrule
English EU DGT  & mDeBERTa-chunk & 0.70 & 0.714 & 0.90 & 0.900 & 0.80 \\
English EU DGT  & XLM-R-chunk    & 0.66 & 0.737 & 0.62 & 0.584 & 0.66 \\
English OpenWHO & mDeBERTa-chunk & 0.69 & 0.670 & 0.80 & 0.802 & 0.74 \\
English OpenWHO & XLM-R-chunk    & 0.83 & 0.755 & 0.90 & 0.965 & 0.86 \\
English SMOL    & mDeBERTa-chunk & 0.75 & 0.812 & 0.76 & 0.702 & 0.76 \\
English SMOL    & XLM-R-chunk    & 0.70 & 0.643 & 0.88 & 0.959 & 0.80 \\
\bottomrule
\end{tabular}
\caption{Per-class precision (P), recall (R), and macro-F1 for chunk-level English encoders. }
\label{tab:prf}
\end{table}

\subsection{Encoder HT Accuracy and Average MT Accuracy}
\label{app:ht-mt}
Tables~\ref{tab:ht-vs-mt-high} and~\ref{tab:ht-vs-mt-low} report per-class HT recall and average MT recall (over translators) for the high- and low-resource datasets, respectively, complementing the balanced-accuracy numbers in Table~\ref{tab:encoder} (averaging the two per row recovers the corresponding balanced accuracy).

\begin{table}[htb]
\centering
\small
\begin{tabular}{llcc}
\toprule
\textbf{Language \& Dataset} & \textbf{Model} & \textbf{HT recall} & \textbf{Avg.\ MT recall} \\
\midrule
English EU DGT   & mDeBERTa        & 0.600 & 0.720 \\
English EU DGT   & mDeBERTa-chunk  & 0.714 & 0.900 \\
English EU DGT   & XLM-R           & 0.597 & 0.833 \\
English EU DGT   & XLM-R-chunk     & 0.737 & 0.584 \\
\midrule
English OpenWHO  & mDeBERTa        & 0.367 & 0.688 \\
English OpenWHO  & mDeBERTa-chunk  & 0.670 & 0.802 \\
English OpenWHO  & XLM-R           & 0.333 & 0.691 \\
English OpenWHO  & XLM-R-chunk     & 0.755 & 0.965 \\
\midrule
English SMOL     & mDeBERTa        & 0.405 & 0.763 \\
English SMOL     & mDeBERTa-chunk  & 0.812 & 0.702 \\
English SMOL     & XLM-R           & 0.357 & 0.724 \\
English SMOL     & XLM-R-chunk     & 0.643 & 0.959 \\
\midrule
Spanish EU DGT   & mDeBERTa        & 0.634 & 0.666 \\
Spanish EU DGT   & mDeBERTa-chunk  & 0.662 & 0.718 \\
Spanish EU DGT   & XLM-R           & 0.632 & 0.648 \\
Spanish EU DGT   & XLM-R-chunk     & 0.774 & 0.766 \\
\midrule
Spanish OpenWHO  & mDeBERTa        & 0.578 & 0.762 \\
Spanish OpenWHO  & mDeBERTa-chunk  & 0.738 & 0.822 \\
Spanish OpenWHO  & XLM-R           & 0.514 & 0.806 \\
Spanish OpenWHO  & XLM-R-chunk     & 0.712 & 0.888 \\
\bottomrule
\end{tabular}
\caption{HT recall and average MT recall (over translators) for high-resource datasets.}
\label{tab:ht-vs-mt-high}
\end{table}

\begin{table}[htb]
\centering
\small
\begin{tabular}{llcc}
\toprule
\textbf{Dataset} & \textbf{Model} & \textbf{HT recall} & \textbf{Avg.\ MT recall} \\
\midrule
Afrikaans FLORES+ & mDeBERTa        & 0.624 & 0.716 \\
Afrikaans FLORES+ & mDeBERTa-chunk  & 0.962 & 0.978 \\
Afrikaans FLORES+ & XLM-R           & 0.612 & 0.668 \\
Afrikaans FLORES+ & XLM-R-chunk     & 0.854 & 0.886 \\
\midrule
Afrikaans SMOL     & mDeBERTa        & 0.626 & 0.714 \\
Afrikaans SMOL     & mDeBERTa-chunk  & 0.932 & 0.948 \\
Afrikaans SMOL     & XLM-R           & 0.722 & 0.778 \\
Afrikaans SMOL     & XLM-R-chunk     & 0.784 & 0.816 \\
\midrule
Swahili FLORES+   & mDeBERTa        & 0.616 & 0.704 \\
Swahili FLORES+   & mDeBERTa-chunk  & 0.712 & 0.748 \\
Swahili FLORES+   & XLM-R           & 0.602 & 0.678 \\
Swahili FLORES+   & XLM-R-chunk     & 0.732 & 0.768 \\
\midrule
Swahili SMOL       & mDeBERTa        & 0.564 & 0.616 \\
Swahili SMOL       & mDeBERTa-chunk  & 0.602 & 0.638 \\
Swahili SMOL       & XLM-R           & 0.522 & 0.578 \\
Swahili SMOL       & XLM-R-chunk     & 0.814 & 0.866 \\
\bottomrule
\end{tabular}
\caption{HT recall and average MT recall (over translators) for low-resource datasets.}
\label{tab:ht-vs-mt-low}
\end{table}

\subsection{Encoder Performance Across Languages and Datasets}
\label{app:encoder-fig}
\begin{figure}[htb]
\begin{center}
\includegraphics[width=0.95\linewidth]{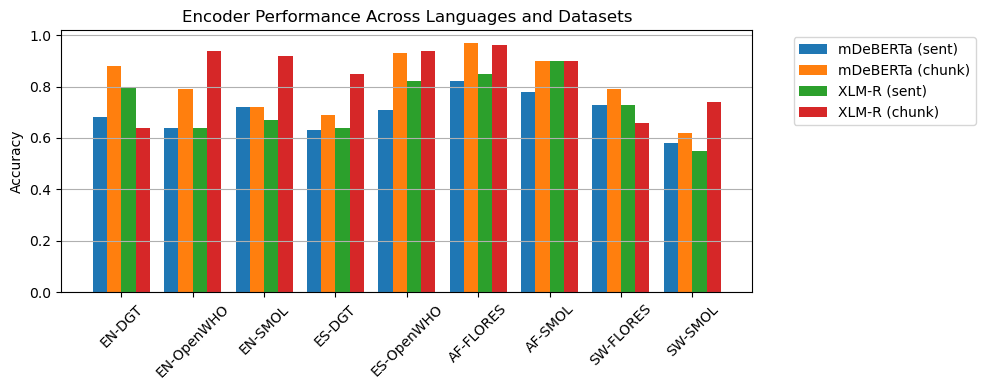}
\end{center}
\caption{Encoder-based MT detection performance across languages and datasets. Results are shown for mDeBERTa and XLM-R under sentence-level and chunk-level (300-token) settings. High-resource results are reported for English and Spanish across EU DGT, OpenWHO, and SMOL; low-resource results for Afrikaans and Swahili across FLORES+ and SMOL. Chunking improves performance consistently for mDeBERTa, whereas the effect is more variable for XLM-R.}
\label{fig:encoder-performance}
\end{figure}

Figure~\ref{fig:encoder-performance} visualizes the balanced-accuracy results of Table~\ref{tab:encoder} across all languages, datasets, and input granularities.

\subsection{Translator-Level Variability in MT Detectability}
\label{app:boxplots}
To better understand variability across translation systems, we provide box plots of MT detection performance computed over individual translators. Unlike aggregate metrics, these plots capture the distribution of detection accuracies across MT systems for each dataset and model.

\begin{figure}[htb]
\centering
\includegraphics[width=0.95\linewidth]{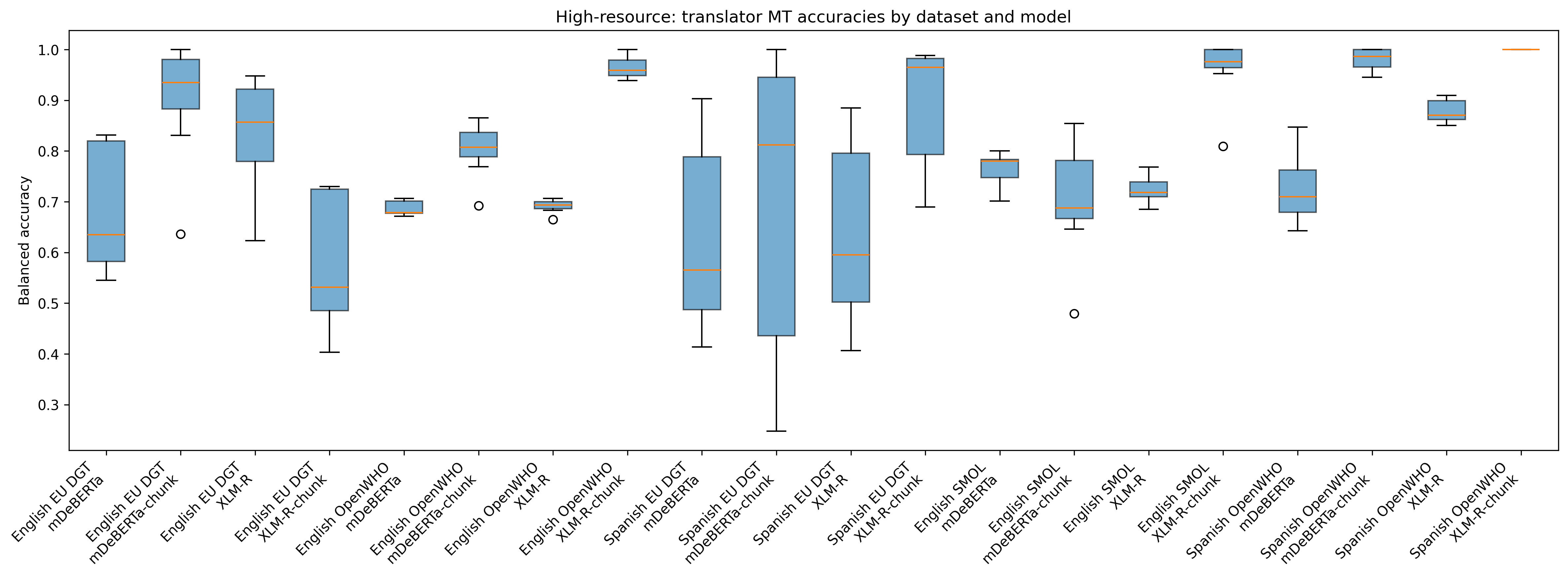}
\caption{Translator-level MT detection accuracies for high-resource datasets. Each box is the distribution of detection performance across MT systems for a given dataset and model; wider boxes indicate greater variability across translators.}
\label{fig:box-high}
\end{figure}

\paragraph{High-resource languages.}
Figure~\ref{fig:box-high} shows translator-level MT detection accuracies for high-resource datasets. We observe substantial variability across translators, particularly in the legislative domain (EU DGT), where interquartile ranges are wide for several models, indicating that detectability depends strongly on the specific MT system. Some settings (e.g., chunked models on OpenWHO) exhibit tighter distributions, suggesting more consistent behavior across translators.

\begin{figure}[htb]
\centering
\includegraphics[width=0.95\linewidth]{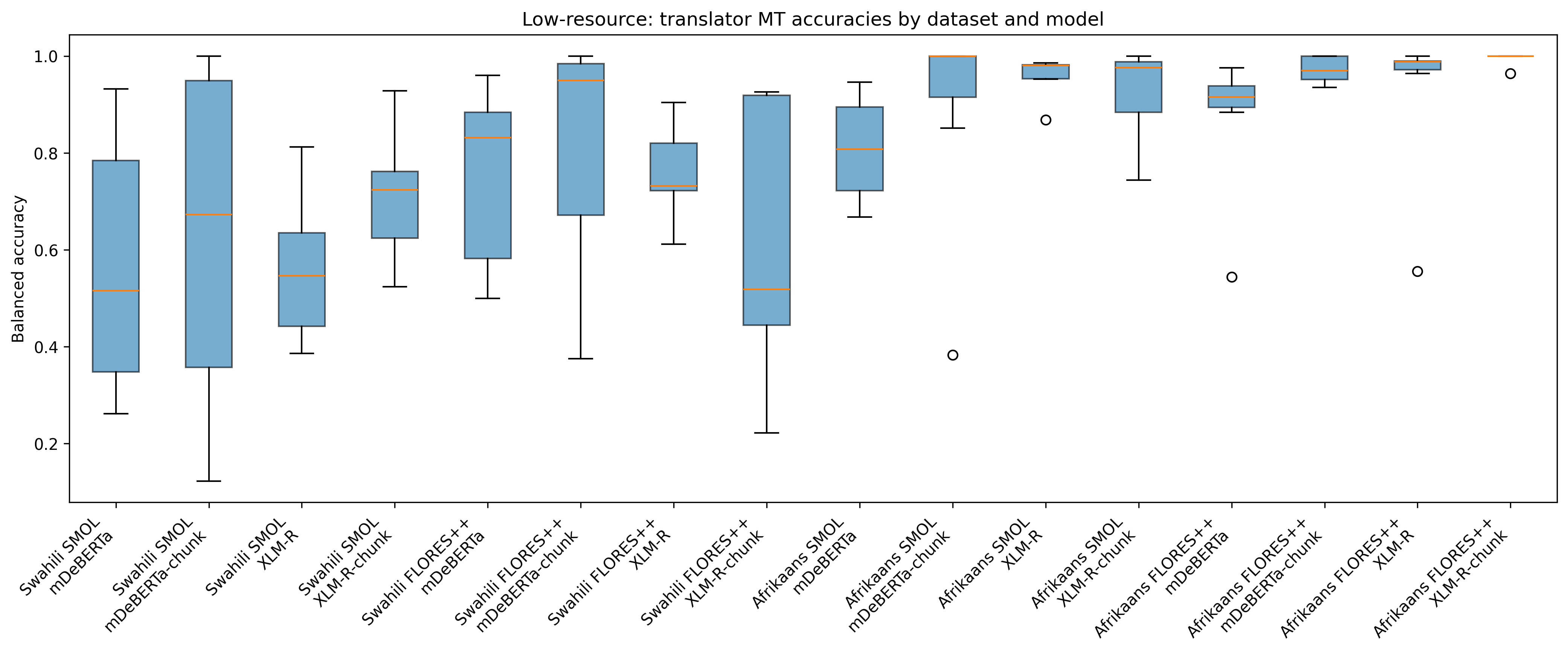}
\caption{Translator-level MT detection accuracies for low-resource datasets. Variability across translators persists despite generally higher detectability.}
\label{fig:box-low}
\end{figure}

\paragraph{Low-resource languages.}
Figure~\ref{fig:box-low} presents the corresponding analysis for low-resource datasets. While overall detection is higher in many cases, variability across translators remains evident, particularly for Swahili. Some models (e.g., encoder-based classifiers) show consistently high detection across translators, while others exhibit wider spreads.

\subsection{Quality-Bin Analysis (COMET and BLEURT)}
\label{app:bleurt}
\begin{figure}[htb]
\begin{center}
\includegraphics[width=0.95\linewidth]{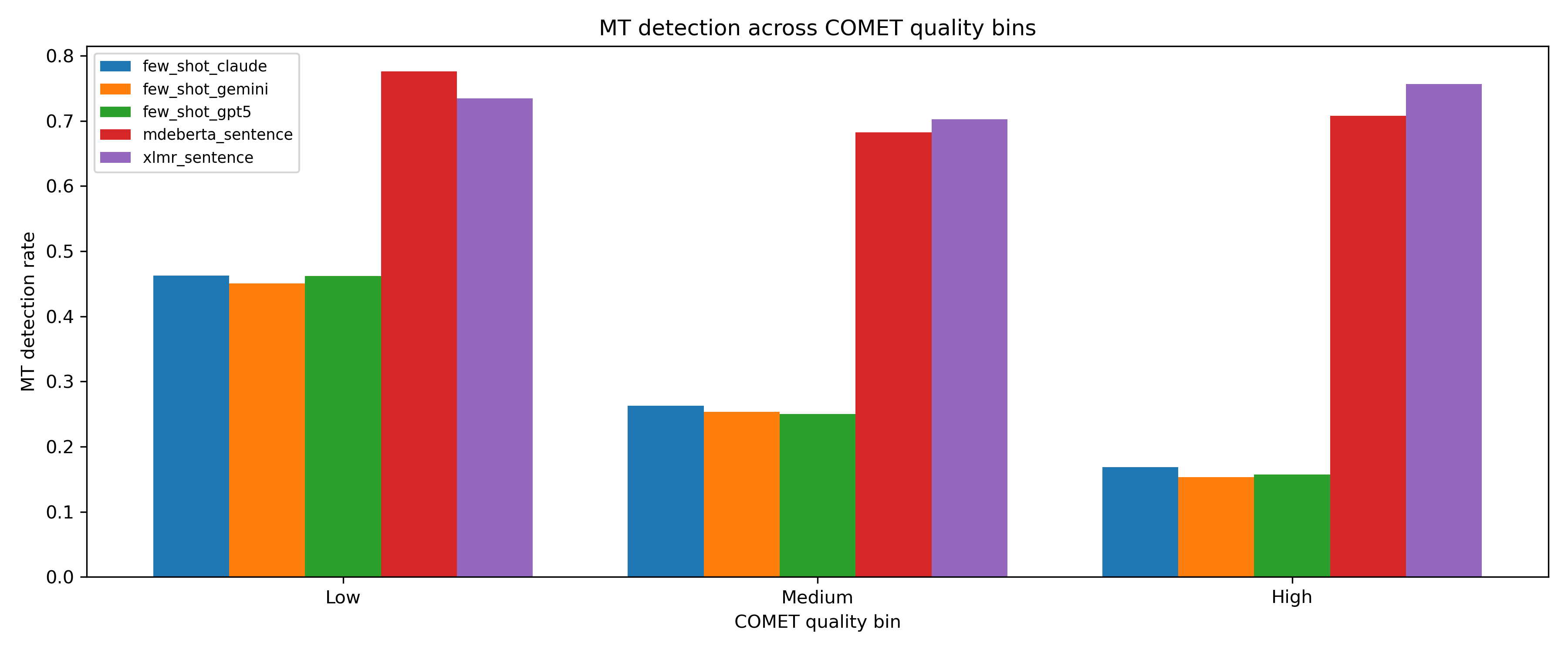}
\end{center}
\caption{MT detection across COMET quality bins. For the few-shot LLMs, detection rates decrease as quality increases; sentence-level encoders remain relatively stable.}
\label{fig:comet-bins}
\end{figure}

\begin{figure}[htb]
\centering
\includegraphics[width=\linewidth]{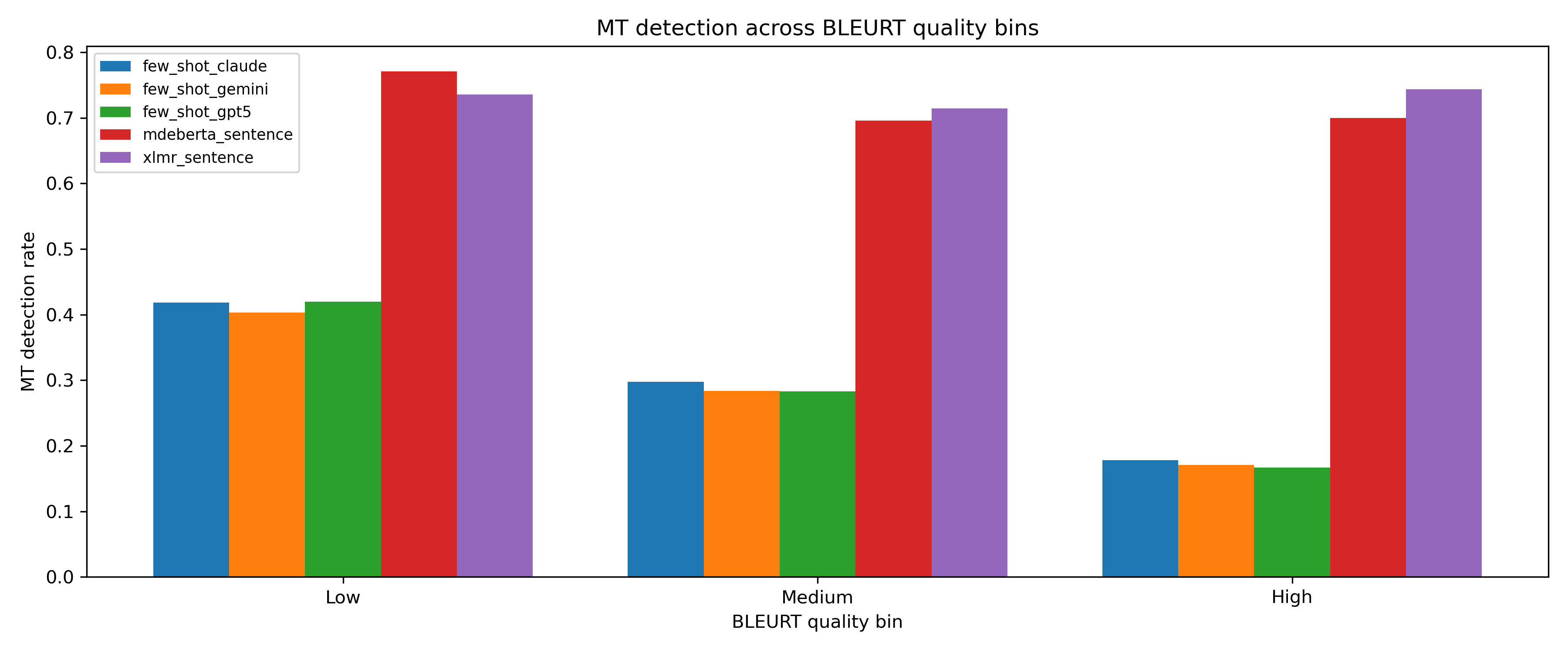}
\caption{MT detection across BLEURT quality bins. Compared to COMET, the relationship between quality and detectability is weaker, although few-shot LLMs still detect lower-quality translations more often than higher-quality ones.}
\label{fig:appendix-bleurt-bins}
\end{figure}

Figure~\ref{fig:comet-bins} visualizes the COMET quality-bin detection rates of Table~\ref{tab:comet-bins}, and Figure~\ref{fig:appendix-bleurt-bins} shows the corresponding BLEURT-based analysis.

\subsection{Translator-Level Quality vs Detectability}
\label{app:translator-quality}
\begin{figure}[htb]
\centering
\includegraphics[width=\linewidth]{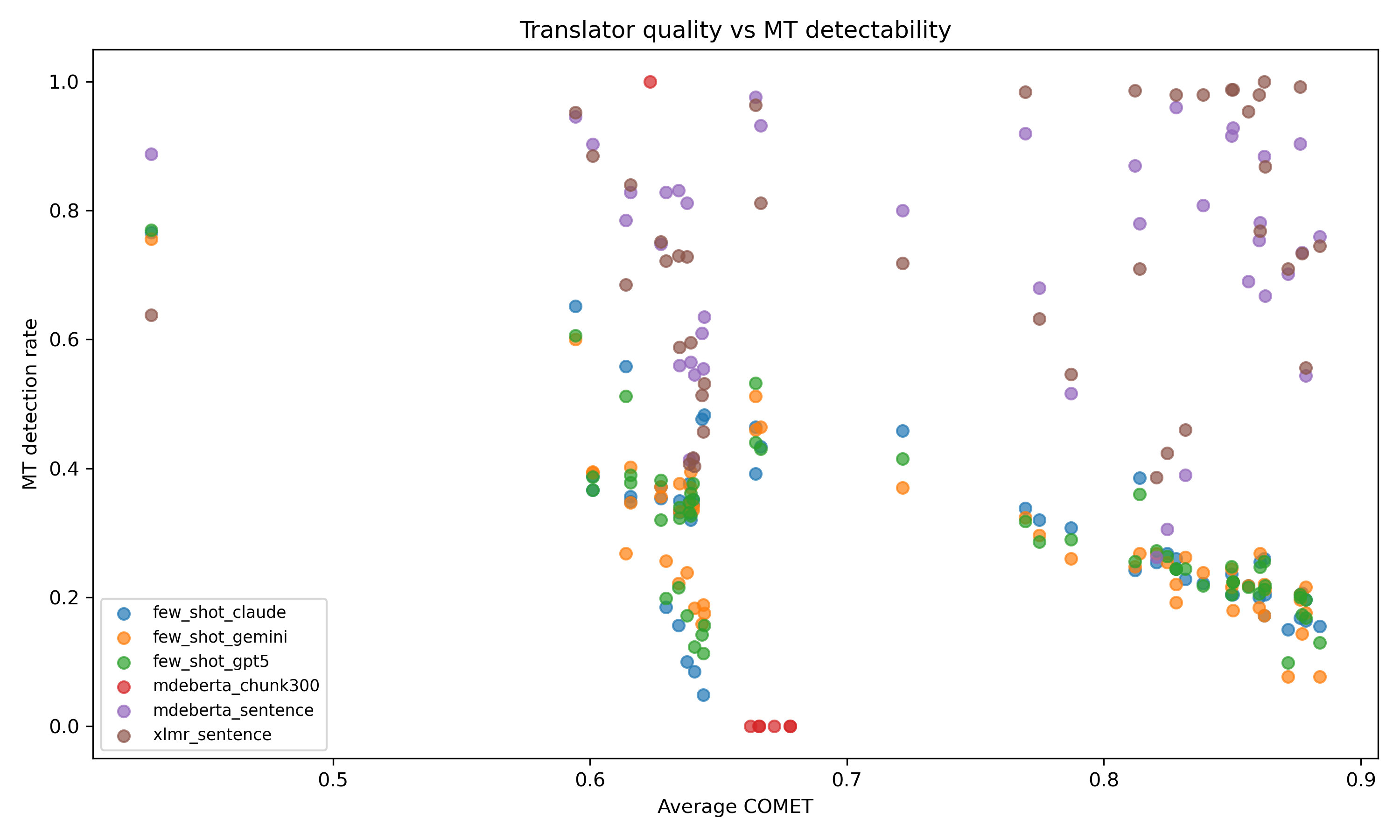}
\caption{Translator-level relationship between average COMET score and MT detection rate across detectors, languages, and datasets. Few-shot LLMs show decreasing detection as translator quality increases, while encoder-based models exhibit weaker, less consistent dependence.}
\label{fig:appendix-translator-comet}
\end{figure}

Figure~\ref{fig:appendix-translator-comet} plots average COMET score against MT detection rate at the translator level, across detectors, languages, and datasets, summarizing the quality--detectability relationship discussed in Section~\ref{sec:discussion}.

\end{document}